\documentclass[11pt,a4paper,oneside,abstracton]{scrartcl}
\usepackage[utf8]{inputenc}
\usepackage[english]{babel} 
\usepackage{amsmath, amssymb, amsthm} 
\usepackage{tabularx, multirow} 
\usepackage{graphicx, tikz} 
\usetikzlibrary{fit,shapes,arrows,arrows.meta,calc,positioning,decorations,backgrounds}
\usepackage{verbatim} 
\usepackage{enumitem} 
\usepackage{color} 
\usepackage{booktabs}
\usepackage{threeparttable}
\usepackage{linguex} 
\usepackage{todonotes} 
\usepackage{hyperref} 
\usepackage{float}
\graphicspath{ {./} }

\KOMAoptions{BCOR=5mm} 
\KOMAoptions{DIV=13}   

\begin{document}
\title{Sense disambiguation of compound constituents}

\author{Carlo Schackow, Stefan Conrad \& Ingo Plag \\[1em]
             Heinrich-Heine-Universität Düsseldorf \\
             40225 Düsseldorf, Germany \\
             \{carlo.schackow,stefan.conrad,ingo.plag\}@hhu.de}
             
\date{\today} 
\maketitle 
\tableofcontents 

\begin{abstract}
In distributional semantic accounts of the meaning of noun-noun compounds (e.g. \textit{starfish, bank account, houseboat}) the important role of constituent polysemy remains largely unaddressed (cf. the meaning of \textit{star} in \textit{starfish} vs. \textit{star cluster} vs. \textit{star athlete}). Instead of semantic vectors that average over the different meanings of a constituent, disambiguated vectors of the constituents would be needed in order to see what these more specific constituent meanings contribute to the meaning of the compound as a whole. This paper presents a novel approach to this specific problem of word sense disambiguation: set expansion. We build on the approach developed by Mahabal et al. \cite{mahabal} which was originally designed to solve the analogy problem. We modified their method in such a way that it can address the  problem of sense disambiguation of compound constituents. The results of experiments with a data set of almost 9000 compounds (LADEC \cite{ladec}) suggest that this approach is successful, yet the success is sensitive to the frequency with which the compounds are attested. 
\end{abstract}

\section{Introduction}

A large part of the literature on the meaning of compounds has focused on the semantic relation between the constituents, both in theoretical linguistics (see \cite[ch. 20]{BLP.2013} for an overview) and psycholinguistics (see, for example, \cite{schmidtke2018conceptual} for recent discussion). Most approaches work on the assumption that each compound constituent represents a particular concept and that these concepts are combined to derive the meaning of the compound. Importantly, however, compound constituents often have more than one meaning, and particular compounds may involve quite different meanings of the same constituent \cite{schafer2020constituent}. Consider, for instance, the data in \ref{ex:poly} where the constituent \textit{chain} has three different readings, depending on in which position it occurs in which compound.

\ex. \label{ex:poly}
\a.	\textit{\textbf{chain}saw}, \textit{bicycle \textbf{chain}}	--- `connected series of metal links'
\b. \textit{\textbf{chain} reaction}	---	`sequence, series'
\c. \textit{supermarket \textbf{chain}} --- `a set of businesses controlled by one firm'

Existing models of compound meaning abstract away from these complications, which is detrimental to an adequate description and understanding of compound semantics. One way of addressing this concern is to disambiguate compound constituents before using the meaning of the constituents to compose the meaning of the compound. There are distributional semantic models that take into account at least the role of the constituent as either modifier or head (e.g. \textsc{caoss}, \cite{MarelliCompoundingAsAbstract.2017}), but the important role of constituent polysemy remains unaddressed. What is needed are disambiguated vectors of the constituents to see what these more specific meanings contribute to the meaning of the compound. 

There are different approaches available in distributional semantics to create disambiguated vectors, e.g. language-based clustering techniques (e.g. \cite{McCarthyWordsense.2016}) or sentence vectors (\cite{SchutzeAutomaticWordSense.1998, LapesaDisambiguation.2018}). The present work attempts to establish set expansion as a method for word sense disambiguation of compound constituents. In particular, the `Category Builder' algorithm, developed by Mahabal et al. \cite{mahabal}, was adapted from the general use case of polysemy to solve the more restricted task of compound constituent disambiguation. The problem of word sense disambiguation is delineated in the next section, followed by an introduction to set expansion.  Then we present Mahabal et al.'s approach, i.e. the Category Builder algorithm,  in section 4. Section 5 comprises the description of our experiments together with the results we obtained.  In addition, we introduce a disambiguation approach based on Wu and Palmer's \cite{wu1994verb} similarity measure to demonstrate that our approach outperforms rather simple approaches.  We conclude with a short summary and discussion.

\section{Word sense disambiguation}

Word sense disambiguation describes the task to identify the senses of a given word as activated by a given context. A context for a word can be a set of other words in the corpus that is somehow related to the target word. For example, the context for a word can be the surrounding words in the sentence the word appears in. The context then induces one or more word senses of the target word. Different contexts may thus activate different senses associated with a given word form. Example \ref{ex:bank} illustrates the classical case of \textit{bank}.

\ex. \label{ex:bank}
\a. ``All my money is in the \textbf{bank}.'' --- `financial institution'
\b. ``The drunken sailor woke up on the \textbf{bank} of the river.'' --- `edge of a watercourse'

In the example above, the two senses of the word \textit{bank} are activated by two different contexts. The word \textit{bank} presents a straightforward example for word sense disambiguation, as the two senses, `financial institution' and `edge of a watercourse', are unrelated, their senses do not overlap. Such cases of clear homonymy are, however, not the only kind of possible constellations of ambiguity. Generally, word senses may overlap partially and completely. In such cases the disambiguation task is expanded by the search for the appropriate precision, with which the word sense is to be chosen. Consider the scientific classification of animals, which supplies examples of word senses that overlap completely, creating a hierarchy of word senses:

\ex. \label{ex:whale}
Cetacea $\longrightarrow$ Balaenopteridae $\longrightarrow$ Balaenoptera $\longrightarrow$ Balaenoptera musculus

In the taxonomy in \ref{ex:whale}, the Latin hyperonym \textit{Cetacea} unifies all the word senses to the right (among many others). The family \textit{Balaenopteridae} again unifies all word senses to its right, and the genus \textit{Balaenoptera} contains the word sense \textit{Balaenoptera musculus}, also known as the Blue Whale. Assuming the whale designated in the context is `Blue Whale', the task is to find the granularity with which the word sense should be chosen. How can such a decision task be implemented?

First, training for such a decision task is word-specific, as the decision is heavily reliant on the semantic encoding of the input word. Specifically, sufficient training data for every sense of the input word is required for training. Secondly, we need to employ supervised learning strategies. This means that handcrafted datasets are needed that are tailored for the specific task. The categories in which to split the input word, as well as the granularity with which to split word senses, need to be chosen before training, as the encoding of the lexicon depends on the intended categories and the envisaged granularity.  In the example of the classification of whales above, we would need to

\begin{itemize}
    \item decide which of the senses are relevant to the task and should be encoded, and then
    \item decide how to encode all the senses into a unique point, or combine some or all of them into clusters.
\end{itemize}

In the next section, we will lay out an architecture that uses set expansion for the disambiguation task.

\section{Set expansion and analogies}

The term `set expansion' is used to refer to the task of expanding a set with suitable related elements to complete the set \cite{Wang2007LanguageIndependentSE}. The usual goal is to find a maximally related word for a given set of input words. The solution is the word in the lexicon with the minimum total distance to all words in the input set. This can be seen as building a category, or class, of words where the input words fit the category, and the solution is the word that best fits the category.
In example \ref{ex:colors}, the input set consists of the words \textit{green} and \textit{yellow}. The solution is added to the set, in this case the  word \textit{red}. Needless to say, the category at issue here is \textsc{Color}. 

\ex. \label{ex:colors}    [green, yellow] $\longrightarrow$ [green, yellow, red]

Set expansion is conceptually similar to word sense disambiguation, in that both tasks utilize categories built up from contexts. 

At the conceptual level, set expansion can also be seen as an approximate opposite of word sense disambiguation. Whereas word sense disambiguation seeks to find all categories that a word fits in, set expansion seeks to find all words that fit in a given category. It is a different approach to categories, and one that allows us to use unsupervised learning strategies in set expansion. In the present paper we use the approach developed by Mahabal and colleagues \cite{mahabal}. In that paper the authors introduce a novel method to solve the problem of finding an analogous word given a context and an alternative context (`analogy problem'). The analogy problem is illustrated in \ref{ex:analogy}:

\ex. \label{ex:analogy} ``What is for the foot what the glove is for the hand?''

In \ref{ex:analogy}, the original word, for which the analogy is to be found, is \textit{glove}, the original context is  \textit{hand}, and the alternative context is \textit{foot}. Note that the solution to a given analogy puzzle does not need to be unique, in this instance \textit{shoe} or \textit{sock} could both be argued to solve the analogy correctly. Mahabal and colleagues use Category Builder to solve a harder variant of the analogy problem: The original context is omitted, leaving only the original word and the new context as elements of the problem:

\ex.    ``What is the glove of the foot?''

To solve the analogy, Category Builder is used to expand the set of [glove] and the set of [foot], and the scores of any words in the union of both expansions are combined. The highest-scoring word in the resulting list solves the analogy.

Category Builder was originally developed with the analogy problem in mind. We adapt this method to solve the problem of sense disambiguation of compound constituents.

\section{The original algorithm of Category Builder}

The algorithm for Category Builder by Mahabal et al. relies on the context-dependent similarity of words. The general idea shared with other work in distributional semantics is that words which appear more often in the same contexts in natural language are more similar to each other. Unlike in other methods, the words are not embedded, i.e. vectorized.  Instead, we hold all words and contexts in sparse incidence matrices, where contexts are in the columns and words are in the rows. The measure used to store the co-incidence of contexts with words is the Asymmetrical Positive Pointwise Mutual Information (APPMI) \cite{mahabal}. This is a variant of Pointwise Mutual Information (PMI), introduced by Church and Hanks \cite{church-hanks-1990-word}:

\begin{equation}
PMI(w, c) = \log \frac{P(w, c)}{P(w) P(c)}
\end{equation}

Here, $ P(w, c) $, $P(w)$ and $P(c)$ are the probability of the coincidence of the word $ w $ and the context $ c $, the probability of the appearance of $ w $ overall, and the probability of the appearance of $ c $ overall, respectively.

PMI can then be used to give a measure for how often a given word appears in a given context, weighted by the likelihood of the context and the word. However, this measure does not take into account the absolute occurrences of words and contexts, only their relative frequency. To include this, we need to amend PMI with APPMI.

For APPMI, a new log term is introduced, and all values are shifted by $k$ and clipped to be zero or higher:

\begin{equation}
APPMI(w, c) = \max(0, PMI(w, c) + \log \frac{P(w, c)}{P(w)}  + k)
\end{equation}

The new log term introduces the frequency of the co-occurrence of a word and a context to the measure. If a given word appears in different contexts with the same PMI, the contexts it appears in with a higher frequency will be scored higher, as that co-occurrence is deemed to be less likely to be coincidental. This helps to reign in the effect of rare contexts.

Shifting the complete term up by $k$ and clipping it to values of zero or higher helps to combat the natural uneven distribution of polysemous words. Lesser known senses of a word will naturally appear in fewer contexts, which means that their PMI will be dwarfed by the more common senses of the word. The factor $k$ decides how much grace to extend to the lesser known senses, by shifting them to be higher than zero. Any terms that were lower than $-k$ before will be zeroed out and not be counted later on. Setting a large amount of occurrences to zero has the side effect of limiting the size of the sparse incidence matrices, and allowing for faster computation.

Two incidence matrices are computed over the complete vocabulary and corpus:

\begin{equation}
M_{c, w}^{V \rightarrow C} = APPMI(w, c)
\end{equation}

\begin{equation}
M_{w, c}^{C \rightarrow V} = APPMI(c, w)
\end{equation}

By way of example, consider the incidence matrix for a small corpus C

\begin{center}
\begin{tabular}{r@{}c@{~}l}
  C & = [ & ``She put all her money in the safe.'',\\
    &     & ``She hid her money in her sock.'',\\
    &     & ``She should spend money on her loan in lieu of snacks.'',\\
    &     & ``She put all her money in her bank in Dallas.'',\\
    &     & ``The bank was closed.'',\\
    &     & ``The bank was open.'',\\
    &     & ``The loan was enough to pay off the house.'',\\
    &     & ``The money was gone.'' ]
\end{tabular}
\end{center}
Contexts can be constructed in a number of ways. For this example, construct a context out of each 3-gram surrounding each word in the Vocabulary \ref{ex:V}:

\ex. \label{ex:V} V = [\textit{money}, \textit{bank}, \textit{loan}]

The sentence ``She put all her money in the safe.'' contains the 3-gram \textit{her money in}, which results in the context \textit{her ... in} with an incidence of 1 for the word \textit{money}. Overall, three contexts are created for the word \textit{money}: \textit{her ... in}, \textit{spend ... on} and \textit{the ... was}. With the setting of k=5, we arrive at the measurements shown in Table~\ref{tab:k}.

\begin{table}[ht]

\begin{center}
\caption{\label{tab:k} $APPMI(w,c)$ measurements for corpus C, $k=5$.}

  \begin{tabular}{@{}r|c@{~}c@{~}c@{~}c@{~}c@{~}c@{~}c@{~}c@{~}c@{~}c@{}}
    & money & bank & loan \\ \hline
    her ... in       & 2.06   & 1.77   & 1.95    \\  \hline
    spend ... on     & 3.81   & 0      & 0       \\  \hline
    the ... was      & 1.73   & 2.48   & 2.19    \\  \hline
\end{tabular}
\end{center}
\end{table}

Let us illustrate the computation of the APPMI values with the co-occurrence of \textit{money} in the context \textit{her ... in}. The word \textit{money} occurs 5 times in the corpus, and the whole vocabulary V amounts to a number of tokens of $N$=10, where \textit{money} appears 5 times, \textit{bank} appears 3 times, and \textit{loan} appears twice. We can calculate P(w) = 5/10 = 0.5. \textit{her ... in} contains 5 incidences from the vocabulary, with \textit{money} occurring three times, so P(c) = 5/10 = 0.5 and P(wc) = 3/5 = 0.6. When we plug these values into APPMI we receive 

\begin{equation}
  APPMI(\text{\emph{money}},\ \text{\emph{her ... in}}) = 
  \max(0, \log { \frac{0.6^2}{0.5^2 * 0.5} } + 5) = 2.06
\end{equation}

Notice that this APPMI value is higher than the values for other words in the context, as \textit{money} appears the most times in this context. The word \textit{money} occurs once each in the contexts \textit{spend ... on} and \textit{the ... was}, but the APPMI value for \textit{spend ... on} is higher, because APPMI favours rare contexts via the added second log-term. The words \textit{money} and \textit{loan} both occur once in the context \textit{the ... was}, but the APPMI value for \textit{loan} is higher, as it appears less frequently overall, and is thus tied closer to the specific context.

Set expansion in Category Builder works as follows. Given the incidence matrices $M^{C \rightarrow V}$ and $M^{V \rightarrow C}$ we define the set expansion vector E:

\begin{equation}
E = M^{C \rightarrow V} W_S M^{V \rightarrow C} S
\end{equation}

where
\begin{itemize}
    \item $W_S$ is a score matrix for the contexts in relation to the input set $S$, and
    \item $S$ is the input set, in the form of an incidence vector over the vocabulary.
\end{itemize}

The score matrix $W_S$ is a diagonal matrix over the contexts. Each diagonal value contains the score $s(c)$ for context $c$:

\begin{equation}
s(c) = f(c)^\rho a(c)
\end{equation}

where
\begin{itemize}
    \item $f(c)$ is the fraction of the input set $S$ that appears in the context,
    \item $a(c)$ is the sum of APPMI values of words in the set,
    \item $\rho$ is the penalty value to influence the importance of $f(c)$. With higher values for $\rho$, the adherence to the whole input set becomes more relevant, thus contexts that only fit part of the input set are penalized.
\end{itemize}

The set expansion vector E is the solution to the set expansion problem. E contains a score for each word in the vocabulary, where the highest-scored words are most likely to complete the set.

Let Table~\ref{tab:set1} illustrate the context matrix $M^{C \rightarrow V}$ over a fictional corpus which provides the vocabulary $V$ and the contexts $C$. For simplicity of the example, we set $M^{V \rightarrow C}$ = $(M^{C \rightarrow V})^T$ having in mind that APPMI is in general asymmetric. The table displays the APPMI value of a given word and a context, where the context now is the part of speech of a word in a sentence. In contrast to the previous example, we here use a completely different kind of context in order to show that the context can essentially be defined in an arbitrary way. For the sake of the example, we use fictitious APPMI values.

\begin{table}[H]
	\begin{center}
		\caption{\label{tab:set1} A toy context matrix.} 
	\begin{tabular}{r|c|c|c|c|c|c}
		 & running & swim & hard & painting & tree & smiling\\
		\hline
		noun & 2 & 3 & 0 & 3 & 4.5 & 0\\
		\hline
		adjective & 2 & 0 & 3 & 0 & 0 & 5\\
		\hline
		verb & 2 & 3 & 0 & 3 & 0 & 4\\
		\hline
		adverb & 0 & 0 & 3 & 0 & 0 & 0
	\end{tabular}
\end{center}
\end{table}

For the input set [running, swim] and a $\rho$ value of 1.0 we can now compute the score matrix $W_S$, as shown in Table~\ref{tab:set2}.
The set expansion vector E, as given in Table~\ref{tab:set3}, can then be derived with the above formula.

\begin{table}[H]
	\begin{center}
		\caption{\label{tab:set2} A toy score matrix.} 	\begin{tabular}{r|c|c|c|c}
		 & noun & adjective & verb & adverb\\
		\hline
		noun & 6 & 0 & 0 & 0\\
		\hline
		adjective & 0 & 3 & 0 & 0\\
		\hline
		verb & 0 & 0 & 6 & 0\\
		\hline
		adverb & 0 & 0 & 0 & 0
	\end{tabular}
\end{center}
\end{table}

\begin{table} [H]
  \begin{center}
    \caption{\label{tab:set3} A toy set expansion vector E (scores ordered descending).}
	\begin{tabular}{r|c|c|c|c|c|c}
		word & swim & painting & smiling & tree & running & hard\\
		\hline
		score & 30.0 & 30.0 & 25.0 & 22.5 & 22.0 & 3.0\\
	\end{tabular}
  \end{center}
\end{table}

The word \textit{painting} is the highest-valued candidate, as it co-occurs in two contexts with both words in the input set, thus \textit{painting} can be used to solve the set expansion. \textit{smiling} scores higher than \textit{tree}, as it co-occurs with \textit{running} in the \textit{adjective} context. If we change the coherence value $\rho$ to 3.0, we receive the scores given in Table~\ref{tab:set4}. 

\begin{table}[H]
	\begin{center}
		\caption{\label{tab:set4} A toy set expansion vector E, $\rho$ = \textbf{3.0} (scores ordered descending).}
		\begin{tabular}{r|c|c|c|c|c|c}
		word & swim & painting & tree & smiling & running & hard\\
		\hline 
        score & 30.0 & 30.0 & 22.5 & 21.25 & 20.5 & 0.75\\
		\end{tabular}
\end{center}
\end{table}

The words \textit{smiling} and \textit{tree} switch places with rising $\rho$, as coherence to all elements of the set becomes more prioritized, and \textit{smiling} only co-occurs with one element of the input set in \textit{adjective}. Having explained the tools, let us now turn to the task of disambiguating compound constituents.

\section{Disambiguating compound constituents}

English binary noun-noun compounds appear in one of three possible spellings:

\begin{itemize}
    \item in open form, i.e. with a space between the two constituents, as in \textit{home phone}
    \item hyphenated, as in \textit{chain-smoker}
    \item in closed form, i.e. as a single orthographic word, as in \textit{campfire}
\end{itemize}

By combining the constituents, a new meaning is created that is more than the combination of their respective semantics. Each constituent provides only a part of its semantic content to the resulting compound, and the compound itself may have semantic properties that cannot be derived from the constituents (see \cite[ch. 20]{BLP.2013} for an overview of the semantics of compounds in English).

Determining the sense with which a constituent enters a compound word is not trivial: For example, \textit{bill} has, among others, the two distinct synonyms \textit{beak} and \textit{banknote}. When part of the compound \textit{hornbill} (the name of a bird species, so called from shape of its bill), only the sense \textit{beak} remains useful. To complicate matters further, \textit{bill} is metonymically used to refer to the bird as a whole, and not to its beak (\textit{pars-pro-toto}). This example shows that the disambiguation of compound constituents is not a trivial task. Similar to sense disambiguation in running texts, the context, in this case the compound itself, determines the correct sense. Ideally, the sense of a given constituent in a compound can be mapped to one of its synonyms, but this may not always be the case. In case of metonymic use of compounds´constituents, or in the case of fully opaque compounds like \textit{hogwash} (meaning `nonsense'), the constituents have not retained a clear relation to their meaning as it can be found outside of the compound.

From the perspective of set expansion, the problem of sense disambiguation of compound constituents can be formulated as in \ref{ex:prob}:

\ex. \label{ex:prob}
What is the probability of a given constituent entering a compound word with one of its synonyms as its sense?

\subsection{Implementation}

The Category Builder algorithm can be used to solve the problem of sense disambiguation in compound constituents in the following way: Starting with a set consisting of a compound and one of its constituents, this set is expanded, and the resulting set is searched for synonyms of the constituent. The highest ranking synonym is the result of the sense disambiguation. This approach necessitates finding the synonyms of the constituent and sorting them by relevance to the compound. The input set could in principle be evaluated for all words in the vocabulary, and compared against all the contexts in the corpus. This is an immense task that can, however, be reduced considerably by evaluating the set only for the synonyms of the constituent, and comparing them against all the contexts that the input set or the synonyms of the constituent appear in. The resulting incidence matrices are of a fraction of their original size, allowing for much faster computation. Additionally, we now only solve the original problem: Sorting the synonyms of the constituent by their relevance with regard to the compound.

In the adapted algorithm, the incidence matrices need to be calculated dynamically for each request. The columns constitute all contexts in which at least one element from the input set appears at least once. The rows constitute all elements from the input set. After set expansion, the compound and the constituent need to be removed from the result set. The highest ranking element of the result set is the candidate for the synonym of the constituent in the compound.


For assessing whether the results of our approach using the Category Builder algorithm are meaningful and of high quality, we generated a baseline approach to compare with.  For this, we built a method assuming that syntactic similarity to the compound can aid in identifying the appropriate synonym for the constituent. For each test item, we queried WordNet\cite{wordnet} applying Wu \& Palmer's distance \cite{wu1994verb} between the compound and each synonym of the constituent to find applicable synonyms for the compound.

\subsection{Test Data Set}
The compound words and their constituents used in this work were extracted from The Large Database of English Compounds (LADEC \cite{ladec}), a dataset of 8952 English compound words, their constituents and a number of annotated measures. The contexts for the experiments were derived from the complete English Wikipedia. The synonyms of the constituents were taken from WordNet \cite{wordnet}, and contexts were created for each occurrence of each compound word, its constituents and all their synonyms. 

\begin{figure}[H]
\centering
\includegraphics[width=150mm]{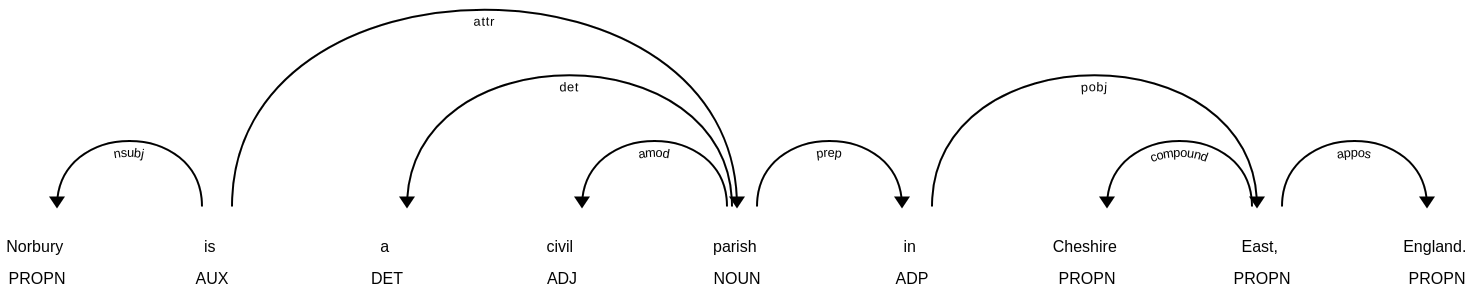}
\caption{Syntactic tree generated for \textit{Norbury is a civil parish in Cheshire East, England}.}
\label{fig:method}
\end{figure}

To obtain the contexts we opted for a syntax-based representation. In a first step, SpaCy \cite{spacy} was used to create an annotated syntax tree for each sentence in the corpus. Each context is then defined as the word's syntactic function in the sentence, the next highest word in the syntactic tree and the word's position in the tree. For illustration, consider the sentence from the corpus \textit{Norbury is a civil parish in Cheshire East, England}. The syntactic tree for this sentence as created by using SpaCy \cite{spacy} is given in Figure~\ref{fig:method}.

When a word from the vocabulary appears in the sentence, a context is generated by concatenating the dependency of the word, its head, and the position of the head (as indicated by its syntactic function). For the word \textit{parish}, the context in this sentence is: [attr, \textit{is}, AUX]. For Cheshire, the context is [compound, \textit{East}, PROPN]. The created contexts are entirely dependent on the use of the word as gauged by its position in the sentence, the word itself is not included in the context.

The test data set was created by hand by one annotator. Each item consists of a compound word, one of its constituents and all synonyms of the constituent. For each pair of compound and constituent, it was decided by the annotator which synonyms are applicable to the compound word. The following two criteria were applied to select viable test items:

\begin{enumerate}
    \item The compound word, the constituent and all the synonyms each appear at least once in the corpus.
    \item The constituent has more than one synonym.
\end{enumerate}

After application of these criteria to all compound words in the LADEC dataset, a test data set of only 213 test items remained. The reason for this massive loss of data is that compound words in the LADEC dataset are rare or outdated, and appear infrequently in the English Wikipedia. Additionally, many constituents of compound words lack appropriate synonyms, or only have appropriate synonyms for the compound word. This makes finding one correct synonym either impossible or trivial.

\subsection{Results}

We started with the baseline approach, i.e. using Wu \& Palmer's distance \cite{wu1994verb} for identifying the best fitting synonym of a constituent in WordNet.  If that synonym was rated by an expert rater as applicable to the compound in the test item, the test item was considered as answered correctly. With this approach, we achieved an accuracy of 43.2 percent. 

Next, we applied our adaptation of the Category Builder to our test data set. For each test item, Category Builder was used to expand the set of the compound and the constituent using the synonyms of the constituent. The highest rated synonym was then compared to the test item. If the synonym was rated as applicable to the compound in the test item, the test item was considered as answered correctly. The algorithm was run once for each of a range of different values of $\rho$. Table~\ref{tab:correct} shows the results.

\begin{table}[ht]
	\begin{center}
	\caption{\label{tab:correct} $\rho$ values and corresponding correctness in the application of Category Builder to the test data set}
    \begin{tabular}{r|c|c|c|c|c|c|c}
        $\rho$     & 1.0  & 1.5  & 2.5  & 3.0  & 3.5  & 4.0  & 5.0 \\
        \hline
        \% of correct results & 56.8 & 59.2 & 58.7 & 58.2 & 58.2 & 58.2 & 58.2 \\
    \end{tabular}
\end{center}
\end{table}

The best parameter setting using this method (with a $\rho$ value of 1.5) resulted in 59.2 percent of correctly answered items. For all values of $\rho$ we obtained at least 56.8 percent of correctly answered items.  This is a promising result, in comparison with our baseline of 43.2 percent.

\begin{figure}[H]
\centering
\includegraphics[width=80mm]{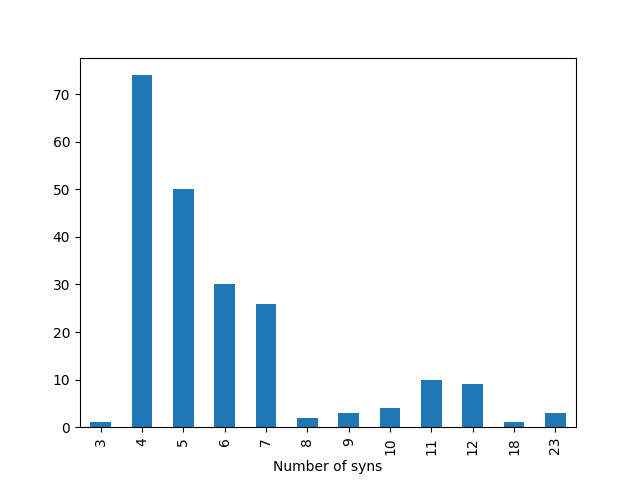}
\caption{The distribution of the number of synonyms over the items in the test data set}
\label{fig:num_syns}
\end{figure}

Every test item had an average of 5.2 synonyms, and 1.6 synonyms were rated as correct by the annotator on average. The distribution of the number of synonyms per item can be found in Figure~\ref{fig:num_syns}.

The mean of 5.2 synonyms with 1.6 correct synonyms among them means that, on average, there was about a 30 percent chance of agreement between annotator and algorithm under random conditions. The Fleiss Kappa value \cite{fleiss1973} for the overall agreement between annotator and algorithm for the test items was $\kappa$ = 0.36. A look at correct vs. incorrect answers is also interesting.

Correctly answered items had an average of 5.6 synonyms and 1.7 synonyms judged as correct by the annotator, compared to 6.7 synonyms and 1.4 synonyms judged as correct for incorrectly answered items. Overall, this signifies fewer options and more correct options on correctly answered items, indicating that the correctly answered items were easier to solve on average then the incorrectly answered items. 

\begin{figure}[ht]
\centering
\includegraphics[width=80mm]{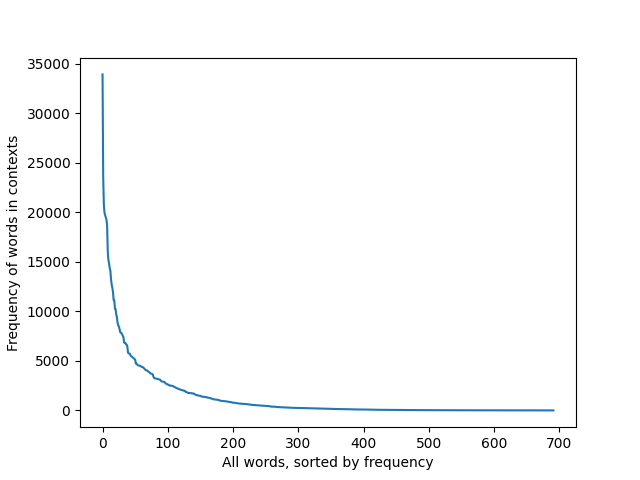}
\caption{Distribution of the number of total occurrences of words used in the test data set}
\label{fig:stims}
\end{figure}

\begin{figure}[ht]
\centering
\includegraphics[width=80mm]{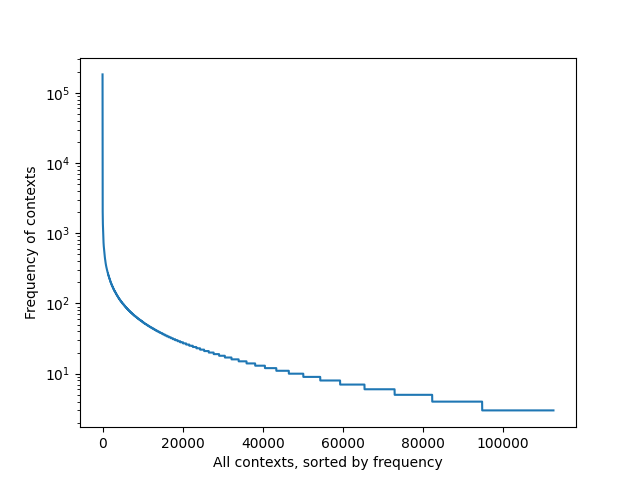}
\caption{Distribution of the number of words per contexts in the test data set}
\label{fig:contexts}
\end{figure}

A more detailed look at the results revealed that correctly answered items appeared in 60 contexts on average. Incorrectly answered items appeared in 25 contexts on average. This suggests that a better data coverage improves the validity of the algorithm. The overall distributions of synonyms and contexts in the test data set can be found in Figures~\ref{fig:stims} and~\ref{fig:contexts}. Both depict the same issue with the data: Most words appear in very few contexts and most contexts hold very few words. 

The $\rho$ value of 1.5 achieved the best results. This value is significantly lower than what is used in the original paper \cite{mahabal} ($\rho$ = 3.0). A high $\rho$ value punishes contexts that contain only part of the input set. The success of a low $\rho$ value suggests that the coherence within the set was not valued highly for this problem. To investigate this further, Category Builder was rerun on the dataset, omitting the constituent in each input set. The percentage of correctly answered items dropped to 33.0 percent. This shows that while coherence in the input set was valued lower than in the original application of Category Builder, the second component could not simply be omitted to simplify the problem. 

\section{Conclusion}

In this paper we have developed a novel approach to the disambiguation of compound constituents. Adapting Mahabal et al.'s Category Builder algorithm, we used set extension to disambiguate constituents based on synonyms. The approach was successful in the sense that it resulted in a much improved accuracy compared to the baseline approach. Many problems remain, however. 

The approach used here relies on the semantic similarity of the compound and its constituents. This is problematic for semantically opaque compounds. Consider, for instance, the word \textit{honeymoon}, where neither \textit{honey}, nor \textit{moon} relate meaningfully to the sense of the compound word. Category Builder is destined to fail on cases like this. Importantly, however, opaque compounds are necessarily outside the scope of any approach that wants to derive compound meaning compositionally.  

The contexts used in the present approach were created using simple syntax trees. The addition of heuristic rules could allow for more complex, information rich contexts and may yield better results. For example, depending on the word function, different word functions with different dependencies could be searched to constitute the context for a word. In Figure~\ref{fig:method}, the context for \textit{parish} was simply constructed by concatenating the dependency, header, and word function of the word: [attr, \textit{is}, AUX]. A more appropriate context could be constructed by searching for adjectives for a noun phrase and the subject for an object phrase, resulting in the two contexts [nsubj, \textit{Norbury}, PROPN] and [amod, \textit{civil}, ADJ]. Different word functions and dependencies would warrant different heuristic rules to be applied. Another goal for future research would be to improve the set expansion approach by refining the process of context creation. 

More detailed comparisons with other disambiguation approaches (like \cite{LapesaDisambiguation.2018,McCarthyWordsense.2016} would be useful to experimentally assess the quantitative (and, thereby, possibly the qualitative) differences between these approaches. Needless to say, the present approach also needs to be tested on other, and preferably larger, data sets to further establish its usefulness for the disambiguation of compound constituents.

\bibliographystyle{abbrv}

\bibliography{main.bib}

\end{document}